\documentclass[twoside,11pt]{article}

\usepackage{blindtext}
\usepackage[symbol]{footmisc}

\usepackage[dvipsnames]{xcolor}
\usepackage{amsmath}
\definecolor{mypink1}{rgb}{0.858, 0.188, 0.478}
\usepackage[preprint]{jmlr2e}

\usepackage{lastpage}

\firstpageno{1}

\begin{document}

\title{Variations and Relaxations of Normalizing Flows } 

\author{\name Keegan Kelly\footnote[2]{All authors contributed equally.} \email kmk9461@nyu.edu \\
       New York University\\
       \AND
       \name Lorena Piedras\footnote[2]{All authors contributed equally.} \email lp2535@nyu.edu \\
       New York University\\
       \AND
       \name Sukrit Rao\footnote[2]{All authors contributed equally.} \email str8775@nyu.edu \\
       New York University\\
       \AND
       \name David Roth\footnote[2]{All authors contributed equally.} \email{dsr331@nyu.edu} \\
       New York University}

\maketitle

\begin{abstract}
Normalizing Flows (NFs) describe a class of models that express a complex target distribution as the composition of a series of bijective transformations over a simpler base distribution. By limiting the space of candidate transformations to diffeomorphisms, NFs enjoy efficient, exact sampling and density evaluation, enabling NFs to flexibly behave as both discriminative and generative models. Their restriction to diffeomorphisms, however, enforces that input, output and all intermediary spaces share the same dimension, limiting their ability to effectively represent target distributions with complex topologies (\citealt{zhang2021}). Additionally, in cases where the prior and target distributions are not homeomorphic, Normalizing Flows can leak mass outside of the support of the target (\citealt{cornish2019, wu2020}). This survey covers a selection of recent works that combine aspects of other generative model classes, such as VAEs and diffusion, and in doing so loosen the strict bijectivity constraints of NFs to achieve a balance of expressivity, training speed, sample efficiency and likelihood tractability. 
\end{abstract}

\color{white}{\footnote[2]{All authors contributed equally.}}
\color{black}

\section{Introduction}

Research in generative modeling with deep learning has in large part focused on four classes of models: flows, VAEs, diffusion models and GANs.
Until recently, GANs had proven the model family capable of producing the highest fidelity generated samples, but a recent string of high-profile successes using diffusion models for natural image (\citealt{ho:20}), audio (\citealt{diffwave}) and video synthesis (\cite{videodiffusion}), trajectory planning (\citealt{janner22}), protein and material design (\citealt{luoantigen, anand2022}) has called into question their dominance in generative tasks. 
VAEs on the other hand, are a slightly older class of models that are easier to train but have been less successful at producing realistic data distributions. 
Some work has gone into improving the expressivity of VAEs (\citealt{aneja2021}) but has encountered a tension between VAE expressivity and a tendency towards posterior collapse, where the generated model ignores the latent codes $z$ entirely in favor of learning a capable generator. 

This paper presents the fundamentals for each of these basic model classes and a selection of recent works that combine aspects from each to achieve a balance of model expressivity, training speed, sample efficiency and likelihood tractability. In particular, we focus on a selection of papers that loosen the strict bijectivity constraints of Normalizing Flows (NF) and attempt to improve the expressivity and sample efficiency of NFs while retaining as much as possible the likelihood evaluation properties the strict construction affords.

\section{Normalizing Flows}
\label{sec:nf}
Normalizing Flows are notable among the broader family of generative models in that they are not only capable of expressing rich, complex distributions-- they are able to do so while also retaining the ability to perform exact density evaluation. They achieve this capacity by expressing a complex target distribution of interest as a bijective, differentiable transformation of a simpler, known base distribution. This formulation provides a learning mechanism using maximum likelihood over i.i.d samples from the target distribution, a sampling mechanism via transformations over points drawn from the base distribution and exact density evaluation using the inverse of the learned transformation and a change of variables with the learned transform's Jacobian.

Normalizing Flows were popularized in the context of Variational Inference by \citet{rezende2015} as a choice of tractable posterior for continuous variables that is more capable of representing complex distributions than traditional choices for approximate posteriors, such as Mean Field Approximations. However, the use of flows for density estimation was first formulated by \cite{tabak2010density} and was used in subsequent works for clustering and classification tasks in addition to density estimation (\citealt{altagnelli2010clustering, laurence2014constrained}).

The formal structure of a Normalizing Flow is as follows: Let $Z \in \mathbb{R}^D$ be a random variable with known probability density function $p_Z$: $\mathbb{R}^D \mapsto \mathbb{R}$, referred to as the base distribution and let $X \in \mathbb{R}^D$ be a random variable of interest over which we would like to define a density $p_X$: $\mathbb{R}^D \mapsto \mathbb{R}$, referred to as the target distribution. We then seek a parameterized transformation $F_{\theta}$: $\mathbb{R}^D \mapsto \mathbb{R}^D$ under which $F_{\theta}(Z) = X.$ We restrict our choices for $F_{\theta}$ to bijective, differentiable mappings, known as \textit{diffeomorphisms}. Under these constraints, the density of a point $x \sim X$, can be calculated under a change of variables using the determinant of the transformation's Jacobian, $J_F$, as follows:
$$p_X(x) = p_Z(z) | det J_F(z) |^{-1}$$
or, framed in terms of the reverse direction,
$$p_X(x) = p_Z(F_{\theta}^{-1}(x)) | det J_F^{-1}(x) |.$$

This product represents the probability density of the inverse-transformed point in the base distribution multiplied by the change in volume incurred by the transformation in an infinitesimal neighborhood around $z$. In practice, $F_{\theta}$ is often constructed as the composition of a sequence of $N$ diffeomorphisms $f_{1,\theta_1}, \ldots, f_{M, \theta_M}$ such that
$$F_{\theta} = f_{1, \theta_1} \circ \cdots \circ f_{M, \theta_M}$$.

Since each of these sub-transformations is itself invertible, their composition is also invertible and bijective. The determinant of $J_F$ can be computed exactly as:
$$det J_F(z) = \prod_{i=1}^N det J_{f_{i, \theta_i}}$$
and the function's inverse as
$$F_{\theta}^{-1} = f_{M, \theta_M}^{-1} \circ \cdots  \circ f_{1, \theta_1}^{-1}.$$


\subsection{Training of Normalizing Flows}
Normalizing Flows can be trained in one of two ways, depending on the nature of access to the target distribution during training. In the setting where samples from $p_x$ are available, but not their densities, the model parameters $\theta$ can be estimated using the forward KL-Divergence:
    \begin{align*}
    \mathcal{L}_{\theta} &=  D_{KL}\left[ p^*_x(x) \mid\mid p_X(x;\theta)\right]\\
    &= - \mathbb{E}_{p^*_x(x)}[\log p_x(x; \theta)] + const.\\
    &= - \mathbb{E}_{p^*_x(x)}[\log p_Z(F_{\theta}^{-1}(x)) + \log | det J_F^{-1}(x) |] + const.\\
\end{align*}
With a set of N samples $\{x_i\}_{i=1}^N$, we can estimate the above loss as
$$\mathcal{L}_{\theta} \approx - \frac{1}{N} \sum_{i=1}^N \log p_Z(F_{\theta}^{-1}(x)) + \log | det J_F^{-1}(x) | + const.$$

In the setting where it is possible to evaluate the target density $p^*_x$ cheaply, but it is not straightforward to draw samples from said distribution, model parameters $\theta$ can be estimated using the reverse KL-Divergence:
\begin{align*}
    \mathcal{L}_{\theta} &=  D_{KL}\left[ p_X(x;\theta) \mid\mid p^*_x(x)  \right]\\
    &= \mathbb{E}_{p_x(x;\theta)}\left[ \log p_x (x;\theta) - \log p_x^*(x) \right]
\end{align*}
\subsection{Limitations of Normalizing Flows}
\label{sec:limitations}
Though Normalizing Flows are in principle capable of representing arbitrarily complex target distributions (\citealt{papamakarios2021normalizing}), for choices of simple base distributions and reasonably smooth transformations they suffer from topological limitations (\citealt{stimper2021}). Strict bijectivity enforces that the input, output and all intermediary spaces share identical dimensionality and topology. \cite{cornish2019} demonstrate that for base and target distributions with distinct support topologies (e.g differing in number of connected components, number of holes), and choice of candidate transformation where $F_\theta$ and $F^{-1}_\theta$ are continuous, it is impossible to represent the target distribution as a transformation of the base distribution and an arbitrarily accurate approximation requires the bi-Lipschitz constant of $F_{\theta}$, a measure of a function's "invertibility" (\citealt{behrmann2020on}) to approach $\infty$. 

\begin{figure}[h]
\centering
\includegraphics[width=0.7\textwidth]{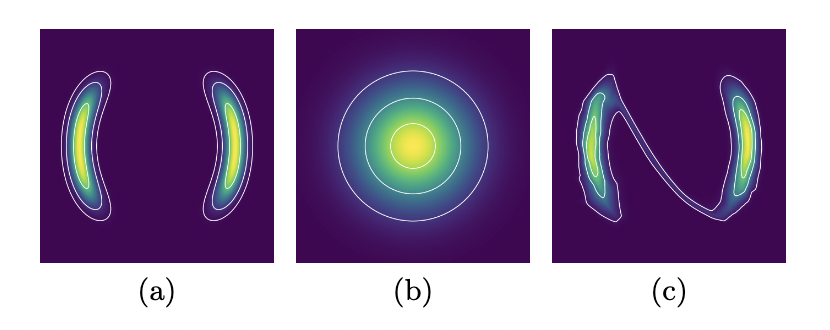}
\caption{An example of "smearing" in ($c$), where the target distribution ($a$) and the base distribution ($b$) differ in their number of connected components.}\textit{Figure taken from \cite{stimper2021}}
\label{fig:smearing}
\end{figure}

Evidence of this limitation can be seen in a "smearing" effect when attempting to represent a bi-modal or multi-modal target distribution using a standard unimodal Gaussian as a base distribution, where sharp boundaries cannot be expressed and density is leaked outside the support of the true target distribution. (\hyperref[fig:smearing]{Figure \ref*{fig:smearing}}) Further, under the \textit{manifold hypothesis} (\citealt{manifoldhypo}), if real-world distributions reside in a low-dimensional  ($r << d$) manifold of the spaces they inhabit, it is a relative certainty that the base and target distributions will have mismatched support topologies and that probability density will leak outside of the target support.

\section{Variational Autoencoders}
\label{sec:vae}
Variational Autoencoders (VAEs) are a likelihood-based class of models that provide a principled framework for optimizing latent variable models (\citealt{kingma-original-vae}).
It consists of two models- a recognition model or encoder and a generative model or decoder that are coupled together. The recognition model approximates the posterior
over the latent random variables which is passed as input to the generative model to generate samples. The generative model, on the other hand, provides a
scaffolding or structure for the recognition model to learn meaningful representations of the data.
The recognition model is the approximate inverse of the generative model according to Bayes' rule. (\citealt{kingma-welling-intro-to-vae})

In the typical setting for a latent variable model,
we have some observed variables and some unobserved variables.
To estimate the unconditional density of the observed variables, also called the model
evidence, we marginalize over the joint distribution of the observed and
unobserved variables, parameterized by $\theta$. This is given by
    \[
      p_{\theta}(x) = \int_{Z} p_{\theta}(x, z)dz
    \]
    
Framing the problem through an implicit distribution over x provides a great deal of flexibility. When we
marginalize over the latents we end up with a compound probability distribution or mixture model.
For example, if $z$ is discrete and $p_{\theta}(x|z)$ is a Gaussian distribution,
then $p_{\theta}(x)$ will be a mixture of Gaussians.
For continuous $z$, $p_{\theta}(x)$, can be seen as an infinite mixture. Thus, depending on the choice of
the latent distribution, we can control the expressivity of the unconditional density $p_{\theta}(x)$, as desired. 

This compound distribution, however, is obtained by integrating over
the support of the latent distribution. Most of the time, this integral is intractable and thus, we cannot differentiate with respect to its parameters and optimize it using gradient descent. While the joint density $p_{\theta}(x, z)$ is efficient to compute, the intractability of $p_{\theta}(x)$, is related to the intractability of the posterior over the latent variable,
$p_{\theta}(z|x)$ (\citealt{kingma-welling-intro-to-vae}).
From the chain rule, we have the following relationship between the densities
\[
  p_{\theta}(z|x)  = \frac{ p_{\theta}(x, z) }{ p_{\theta}(x) }
\]

The intractability of $p_{\theta}(z|x)$ leads to the intractability of  $p_{\theta}(x)$. To overcome this hurdle,
we employ approximate inference techniques. The framework of VAEs
provides a computationally efficient way for optimizing latent variable models jointly with a corresponding inference model
using gradient descent (\citealt{kingma-welling-intro-to-vae}).
This is achieved by introducing the encoder or recognition model- a parametric inference model $q_{\phi}(z|x)$, where $\phi$ is the set of variational parameters. 

\begin{figure}[h]
\centering
\includegraphics[width=0.3\textwidth]{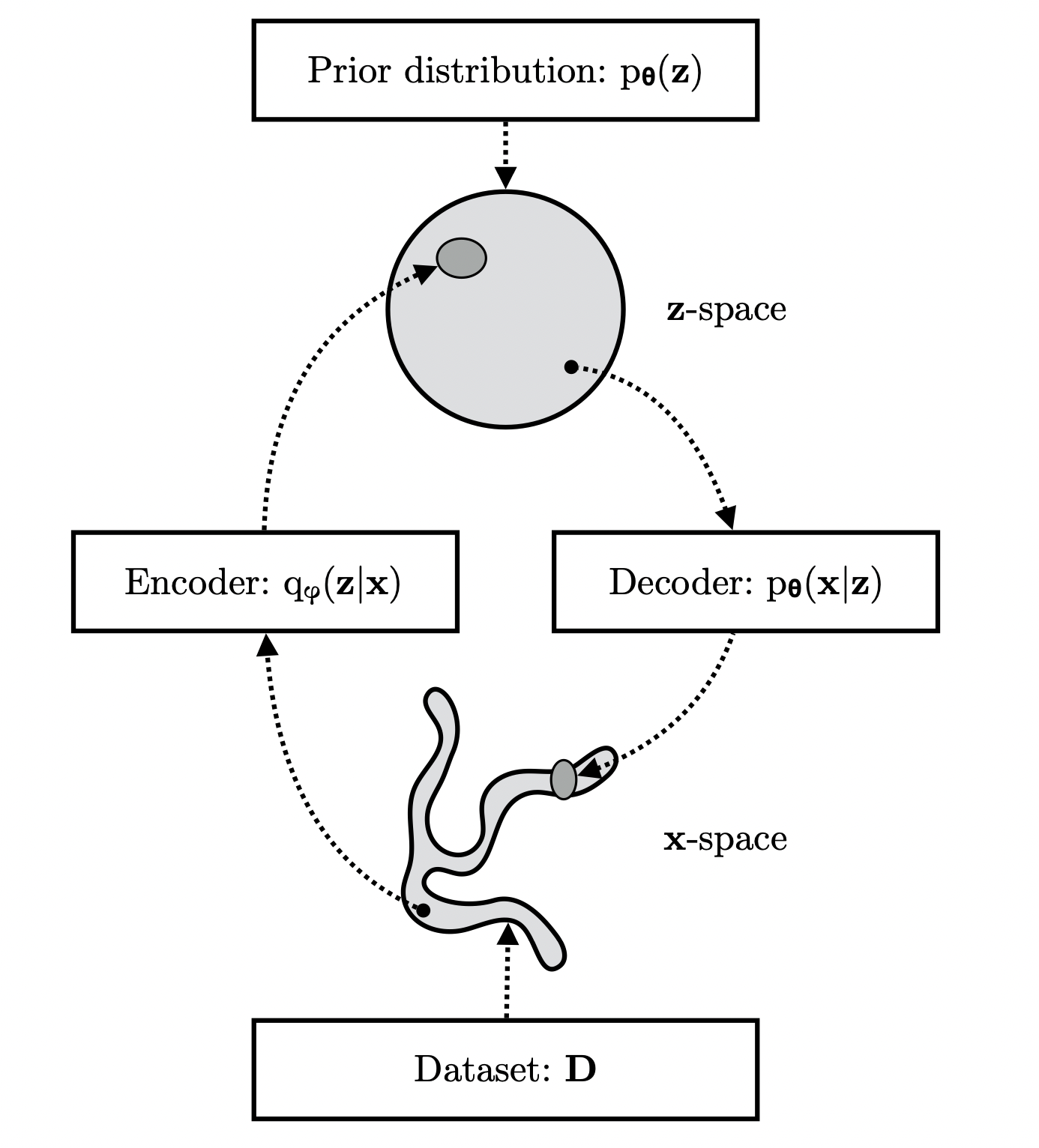}
\caption{Computational flow in a VAE}
\textit{Figure taken from \cite{kingma-welling-intro-to-vae}}
\label{fig:vae-computation-flow}
\end{figure}

Consequently, the optimization objective of VAEs is the variational lower bound or evidence lower bound (ELBO), where we optimize the variational parameters $\phi$ such that
\[
  q_{\phi}(z|x) \approx  p_{\theta}(z|x)
\]
It follows from the derivation shown below
\begin{gather*}
\begin{split}
    \log p_{\theta}(x) &=  \mathbb{E}_{q_{\phi}(z|x)} \log p_{\theta}(x) \\
    \phantom{\log p_{\theta}(x)} &= \mathbb{E}_{q_{\phi}(z|x)} \log \left[ \frac{p_{\theta}(x, z)}{p_{\theta}(z|x)} \right] \quad \quad \quad \quad \quad  \quad \quad \quad \quad \quad \quad  \text{(chain rule)}\\
    \phantom{\log p_{\theta}(x)} &= \mathbb{E}_{q_{\phi}(z|x)} \log \left[ \frac{p_{\theta}(x, z) q_{\phi}(z|x) }{q_{\phi}(z|x)  p_{\theta}(z|x)} \right]\\
    \phantom{\log p_{\theta}(x)} &= \underbrace{\mathbb{E}_{q_{\phi}(z|x)} \log \left[ \frac{p_{\theta}(x, z)}{q_{\phi}(z|x)} \right]}_{\substack{=\mathcal{L}_{\phi, \theta}(x) \\ \text{\ \ (ELBO)} } }
            + \underbrace{\mathbb{E}_{q_{\phi}(z|x)} \log \left[ \frac{q_{\phi}(z|x) }{p_{\theta}(z|x)} \right]}_{=D_{KL}(q_{\phi}(z|x) || p_{\theta}(z|x) )  } \\
\end{split}
\end{gather*}

The second term is the Kullback-Leibler (KL) divergence between $q_{\phi}(z|x)$ and  $p_{\theta}(z|x)$, while the first term is the
variational lower bound or evidence lower bound (ELBO). 

Since the KL divergence is non-negative, the ELBO is the lower bound on the log-likelihood of the data
\begin{gather*}
    \mathcal{L}_{\phi, \theta}(x)  = \log p_{\theta}(x) - D_{KL}(q_{\phi}(z|x) || p_{\theta}(z|x) ) \\
    \mathcal{L}_{\phi, \theta}(x)  \leq \log p_{\theta}(x) \phantom{- D_{KL}(q_{\phi}(z|x) || p_{\theta}(z|x) )} 
\end{gather*}

Thus, we can observe that maximizing the ELBO $\mathcal{L}_{\phi, \theta}(x)$ with respect to $\theta \mbox{ and } \phi$, will have the following consequences
\begin{itemize}
    \item It will approximately maximize the marginal likelihood $p_{\theta}(x)$, implying that our generative model will get better
    \item It will minimize the KL divergence between $q_{\phi}(z|x)$ and  $p_{\theta}(z|x)$, implying our approximation of the posterior, $q_{\phi}(z|x)$, will get better
\end{itemize}

\section{Denoising Diffusion}
\label{sec:ddpm}
Diffusion-based generative models are parameterized Markov chains mainly used to create high quality images and videos, and also utilized in data compression and representation learning on unlabeled data. Diffusion models are both tractable and flexible, making them easier to train, evaluate and sample from. The transitions of the Markov chain gradually add noise to the data and then learn to reverse the diffusion process, producing desired samples after a finite time. Unlike VAEs, diffusion models are latent variable models where the dimensionality of latent space is the same as the original data. The idea of using diffusion for a generative process was initially introduced by \cite{sohl:15} in 2015. \cite{wallach:19} and \cite{ho:20} improved the initial approach a couple of years after. The latter showed that diffusion models were capable of generating high quality images and unveiled an equivalence with denoising score matching in training and Langevin dynamics at the sampling stage. 

The forward diffusion process gradually adds a small amount of noise in $T$ steps to the original data until it is indistinguishable. A variance schedule $\beta_1, \dots, \beta_T$, where $w_i \in (0,1)$,  is used to regulate the size of each step. If the noise is small enough, the transitions of the reverse diffusion process will be conditional Gaussians as well. Given a point sampled from the original data distribution $x_0 \sim q(x)$ we have the following transition probabilities \cite{weng:21}:
\[
q(x_t|x_{t-1}) = \mathcal{N}(x_t;\sqrt{1-\beta_t}x_{t-1}, \beta_t I)
\]
The forward process in a diffusion probabilistic model is fixed, other diffusion models such as diffusion normalizing flows have a trainable forward process \cite{zhang2021}. 
A desirable property of the forward process shown by \cite{sohl:15} is that we can sample $x_t$ given $x_0$ at any time step without having to apply $q$ repeatedly
\[
q(x_t|x_0)=\mathcal{N}(x_t;\sqrt{\bar{\alpha}_t}x_0, (1-\bar{\alpha}_t)I)
\]
with $\alpha_t := 1-\beta_t$ and $\bar{\alpha}_t := \prod_{s=1}^{t}\alpha_s$.

We could use $q(x_{t-1}|x_t)$ to revert the forward diffusion process and generate a sample from the real distribution using random noise as input. Unfortunately  $q(x_{t-1}|x_t)$ is intractable, therefore we will learn a model $p_{\theta}(x_{t-1}|x_t)$ to approximate it. Notably, the reverse conditional probability is tractable if we condition on $x_0$ \cite{weng:21}. Similar to VAE, we can use a variational lower bound to optimize $-\log p_{\theta}(x_0)$. After rewriting the lower bound into several KL-divergence terms and one entropy term and ignoring all terms that don't have any learnable parameters, we get two components $L_0 = -\log p_{\theta}(x_0|x_1)$ and $L_t=D_{KL}(q(x_t|x_{t+1},x_0) || p_{\theta}(x_t|x_{t+1}))$. $L_t$ is the KL divergence of two Gaussian distributions, where $q(x_t|x_{t+1},x_0)$ is the tractable reverse conditional distribution mentioned earlier. In \cite{ho:20}, $L_0$ is modeled using a separate discrete decoder and a fixed variance term. 



\section{Score-Based Methods}
\label{sec:score}
Transport models employ maximum likelihood estimation to learn probability distributions. This reliance can pose a major challenge given complex partition functions, which may prove intractable. Some models add constraints to ensure MLE is feasible; the bijectivity of Normalizing Flows and the approximation of the partition function in VAEs are two such methods to overcome intractability. Another framework to address this scenario is known as the score-based method. For this setup, we model the score function rather than the density function directly: 
$$  s_{\theta}(x) = \nabla_x \log p_{\theta}(x) $$ 

Partition function $Z_{\theta}$ is reduced to zero in the context of $\nabla_x \log Z_{\theta}$  and thus is independent of the score-based model. We are therefore able to sidestep any challenging computation posed by the partition function while training. This setup introduces flexibility, where we can now work with many families of models that may have been otherwise intractable.  

Score-based diffusion is an extension upon this method. As in the previous section on diffusion, this model class involves both a forward and backward stochastic differential equation. Again, the forward pass returns a noisy distribution: 

$$x_t = e^{-t}x + \sqrt{1-e^{-t} }z$$ 

Where $x \sim \pi_d$ and $z \sim N(0, I)$.  

In score-based diffusion, the reversed pass can now be written as a flow composed of diffusion drift plus an exact score: 

$$dY_t = [Y_t - \nabla log \pi_t(Y_t)]dt + \sqrt{2} dB_t$$ 


 
Where $Y_t = X_{T-t}$ (\cite{classnotes}).  

The challenge now falls on estimating the scores from the data. This is particularly impactful in low density regions, where there is less data available to compute scores. In such cases, the model may produce poor quality sampling. To work around this obstacle, noise can be added to the data in increasingly larger magnitudes so that the model can at first train on less corrupted data, then learn low data density regions as the magnitude of noise grows. In this way, adding noise adds stability to score-based methods and aids in producing higher quality samples (\citealt{wallach:19}). 

Score-based models can also be used to compute exact likelihoods. This requires converting the reverse stochastic differential equation into an ordinary differential equation, as seen below:

$$dx = [f(x,t) - \frac{1}{2} g^2(t) \nabla_x \log p_t(x)] dt$$. 

The above equation, known as the probability flow ODE, becomes a representation of a neural ODE when the score function $\nabla_x \log p_t(x)$ is approximated by the score-based model $s_{\theta}(x,t)$. Because of this relationship, the probability flow ODE takes on characteristics of a neural ODE, such as invertibility, and can compute exact likelihoods using the instantaneous change-of-variables formula (\citealt{score:likelihood}).

\section{Relaxing Constraints}
\label{sec:relaxing}

In this section, we explore several works that formulate new model classes by relaxing the strict bijectivity constraints of Normalizing Flows. These works expand the family of admissible functions to include surjective/stochastic transformations and take inspiration from score-based models and diffusion by introducing noise into the training process. 

\subsection{SurVAE Flows}
In an attempt to place VAEs and Normalizing Flows in a common context, \cite{nielsen2020} introduce SurVAE Flows-- a class of models composed of \textit{surjective transformations}, allowing for models that mix bijective, surjective and stochastic components in a single end-to-end trainable framework. They identify three mechanisms needed for probabilistic generative models in this family:
\begin{enumerate}
    \item A forward transformation: $p(x | z)$
    \item An inverse transformation: $p(z | x)$
    \item A likelihood contribution: $p(x)$
\end{enumerate}

In a normalizing flow, the forward transformation and reverse transformations are deterministic and can be represented as $p(x | z) = \delta (x - F(z))$ and $p(x | z) = \delta (z - F^{-1}(x))$. In a VAE, both directions are stochastic and a variational approximation $q(z | x)$ is used in place of the intractable posterior.

They use this decomposition to draw formal connections between stochastic transformations (VAEs) and bijections (normalizing flows) using Dirac delta functions. In particular, they show that the marginal density $p(x)$ can be expressed under both paradigms as:
$$\log p(x) \simeq \log p(z) + \mathcal{V}(x, z) + \mathcal{E}(x,z)$$ where $\mathcal{V}(x, z)$ represents the likelihood contribution and $\mathcal{E}(x,z)$ represents the 'looseness' of the provided bound. Under VAEs and other stochastic transformations, the likelihood contribution term is calculated as $\log \frac{p(x \mid z)}{q(z \mid x)}$ and the 'bound looseness' term is calculated as $\log \frac{q(z | x)}{p(z | x)}$, while under normalizing flows and other bijections, the likelihood contribution term is $\log | \det J |$ and the 'bound looseness' term is $0$.

Through the use of surjective, non-injective layers, the authors present constructions that allow for \textit{inference surjections}-- models with exact inverses and stochastic forward transformations-- and \textit{generative surjections}-- models with exact forward transformations and stochastic right inverses. In doing so, they formulate models that bypass the dimensionality constraints enforced by bijectivity without sacrificing the ability to perform exact likelihood evaluation. 

The surjective layers they introduce include absolute value, max-value and stochastic permutation, which they use to demonstrate strong experimental results on synthetic modeling tasks. They demonstrate the effectiveness of surjective layers on a handful of synthetic modeling tasks, particularly those with inherent symmetries. Importantly for this survey, these experiments also demonstrate an ability to model sharper boundaries than a fully bijective flow is capable of producing.

The authors argue that a number of recently proposed generative model types can be understood as SurVAE flows, including diffusion models (\cite{ho:20}), continuously indexed normalizing flows (\cite{cornish2019}), stochastic normalizing flows (\cite{wu2020}) and normalizing flows acting on augmented data spaces (\cite{huang2020}).

\subsection{Stochastic Normalizing Flows}
\begin{figure}[h]
\centering
\includegraphics[width=0.9\textwidth]{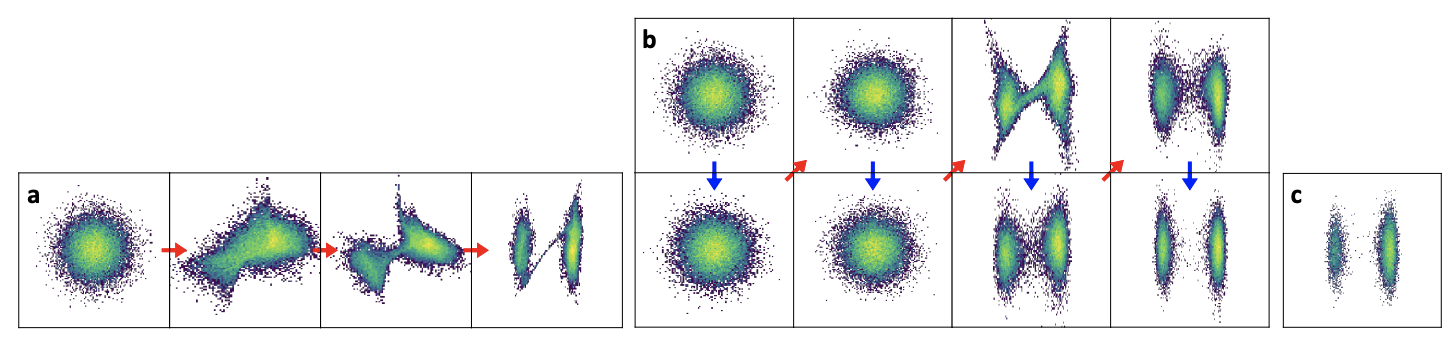}
\caption{Double well problem: a) Normalizing flows, b) NF with stochasticity, c) Sample from true distribution}
\textit{Figure taken from \cite{wu2020}}
\label{fig:double-well-stochastic}
\end{figure}

Stochastic Normalizing Flows (SNF) are a generalization of the Normalizing Flow framework introduced by \cite{wu2020}. They offer certain benefits over classical stochastic sampling methods and Normalizing Flows for sampling from a known energy model specified up to a normalization constant. Sampling methods such as Markov Chain Monte Carlo (MCMC) or Langevin Dynamics (LD) may have trouble converging because of slow mixing times and local energy minima-- adding a deterministic transformation can help alleviate this problem. On the other hand, introducing noise to relax Normalizing Flow's bijectivity constraints can help solve the topological constraints mentioned in \ref{sec:limitations}. In Figure \ref{fig:double-well-stochastic} we show the double-well example, by adding stochasticity we're able to successfully separate the modes of the distributions avoiding the "smearing" effect.

Similar to NFs, SNFs are a sequence of deterministic transformations. Their contribution comes from adding stochastic blocks, such as Langevin, Metropolis-Hastings, VAE, and diffusion normalizing flow layers. Both the deterministic and stochastic transformations help modify a prior into a complicated target distribution. We can use KL divergence to train NFs and SNFs. In the former we can calculate the probability density to generate a sample $p_x(x)$ using change of variables, however, we can no longer do so-- with the introduction of stochasticity, SNFs are no  longer invertible. As described in \ref{sec:nf}, we can train a Normalizing Flow by energy-based training, used when we have a model for the target energy, or maximum likelihood training, when we have samples. We need to generalize the notion of energy and maximum likelihood training in order to train an SNF. We start by defining  $\mu_{z}(x) \propto e^{-u_z(z)}$ as our latent space distribution, $p_x(x) \propto e^{-u_x(x)}$ as our output distribution, and maximizing the importance weights
$$
\log w(z \rightarrow x) = \log{\frac{\mu_x(x)}{p_x(x)}} = -u_x(x) + u_z(z) + \sum_{t}\nabla S_t(y_t)$$
where $y_{t+1}|y_t \sim q_t(y_{t} \rightarrow y_{t+1}),\ y_t|y_{t+1} \sim \Tilde{q}_t(y_{t+1} \rightarrow y_t)$ are the forward/backward probability distributions at $t$, we no longer have deterministic flow layers, and $\nabla S_t = \log \frac{\Tilde{q}_t(y_{t+1} \rightarrow y_t)}{q_t(y_{t} \rightarrow y_{t+1})}$ represents the forward-backward probability ratio of step t. By maximizing the importance weights we get the following expressions for energy base training
$$
\min \mathbb{E}_{\mu_{z}(x) \mathbb{P}_f(z \rightarrow x)}[-\log(w(z \rightarrow x))] = \text{KL}(\mu_{z}(x) \mathbb{P}_f(z \rightarrow x) || \mu_{x}(x) \mathbb{P}_b(x \rightarrow z))
$$
and for maximum likelihood training
$$
\min \mathbb{E}_{\mu_{x}(x) \mathbb{P}_b(x \rightarrow z)}[-\log(w(z \rightarrow x))] = \text{KL}(\mu_{x}(x) \mathbb{P}_b(x \rightarrow z) || \mu_{z}(x) \mathbb{P}_f(z \rightarrow x)).
$$
where $\mu_{z}(x) \mathbb{P}_f(z \rightarrow x),\ \mu_{x}(x) \mathbb{P}_b(x \rightarrow z)$ are our forward and backward pass probabilities. Notably, the KL divergence of the paths is an upper bound to the KL divergence of the marginal distributions.
$$
\text{KL}(p_x||\mu_x) \leq \text{KL}(\mu_{z}(x) \mathbb{P}_f(z \rightarrow x) || \mu_{x}(x) \mathbb{P}_b(x \rightarrow z))
$$
Finally, we can draw asymptotically unbiased samples from our target distribution $x \sim \mu_x(x)$ by employing the Metropolis-Hastings algorithm and using the importance weights shown above.

\subsection{Diffusion Normalizing Flow}

Diffusion Normalizing Flow (\cite{zhang2021}), or DiffFlow, was introduced as a cross between Normalizing Flows and Diffusion models. DiffFlow is composed of two neural stochastic differential equations (SDEs): a forward pass $F$ that transforms the data X into a simple distribution like Gaussian and a backward pass $B$ that removes noise from the simple distribution to generate samples that match the target data distribution. Like diffusion models, the SDEs are jointly trained to minimize the KL divergence between the forward pass distribution and the backward pass distribution at final time $\tau$. The objective equation is as follows:
$$KL(p_{F(\tau)}| p_{B(\tau)}) = \mathbb{E}_{\tau\sim p_F}[\log p_F(x_0)] + \mathbb{E}_{\tau\sim p_F}[-\log p_B(x_N)] + \sum_{i=1}^{N-1} \mathbb{E}_{\tau\sim p_F}[-\log \frac{p_F(x_i|x_{i-1})}{p_B(x_{i-1}|x_{i})}]$$

Similar to Normalizing Flows, DiffFlow is able to learn while mapping to the latent space. However, DiffFlow relaxes the bijectivity constraint of NFs on this mapping. In doing so, Difflow has more expressivity and can learn distributions with sharper boundaries. Further, bijectivity may prevent models from having density support over the whole space. Thus in lifting the constraint, DiffFlow has been found to perform better on tasks like image generation of complex patterns. The authors also claim that the boosted expressivity of DiffFlow results in better performance in likelihood over other NF implementations(\cite{zhang2021}).

Diffusion Normalizing Flow bypasses the bijectivity constraint by adding noise to the forward stochastic differential equation. Most diffusion models add noise indiscriminately, which can require many iterations to reach Gaussian noise and can lead to generated distributions with corrupted or missing details. On the other hand, due to the trainability of the forward SDE, DiffFlow adds noise only to targeted areas. Thus, DiffFlow can diffuse noise more efficiently and retain topological details that might have been blurred out in other diffusion processes.  

Similar to diffusion, DiffFlow SDEs are composed of a drift term f, a vector valued function, and a diffusion term g, a scalar valued function. The equations are as follows: 
\begin{align*}
    \text{Forward SDE:} \quad  dx &= f(x, t, \theta)dt + g(t)dw\\
    \text{Backward SDE:}  \quad dx &= [f(x, t, \theta) - g^2 (t)s(x, t, \theta)]dt + g(t)dw, 
\end{align*}
where $x$ is data at time $t$ and $w$ represents the standard Brownian motion. The main distinguishing factor from Diffusion models is that DiffFlow includes the $\theta$ parameter in the drift term which makes the SDEs learnable. From these equations, it is clear that when the diffusion term $g$ tends to $0$, DiffFlow reduces to Normalizing Flows.  

Given the SDEs above, the discretized equations can be written as:
\begin{align*}
    x_{i+1} &= x_i +f_i(x_i) \Delta t_i + g_i \delta_i^F \sqrt{\Delta t_i}\\
    x_{i} &= x_{i+1} - [f_{i+1}(x_{i+1}) - g^2_{i+1}(x_{i+1})]\Delta t_i + g_{i+1} \delta_i^B \sqrt{\Delta t_i}
\end{align*}

Returning to KL divergence, given that the first term is a constant and utilizing the discretized SDEs, the objective can be reduced to the form:

$$L = \mathbb{E}_{\delta^F; x_0\sim p_0}[-\log p_B(x_N) + \sum_{i=1}\frac{1}{2}(\delta_i^B(\tau))^2]$$

where noise is represented as:
$$\delta_i^B(\tau) = \frac{1}{g_{i+1}\sqrt{\Delta t}} [x_i - x_{i+1} + [f_{i+1}(x_{i+1}) = g^2_{i+1}s_{i+1}(x_{i+1})]\Delta t ] $$.

Loss can now be minimized with Monte Carlo gradient descent (\cite{zhang2021}). 


\subsection{Stochastic Normalizing Flows and Diffusion Normalizing Flows}
\cite{zhang2021} introduced Diffusion Normalizing Flows (DNF) as a new type of model. Nevertheless, per \cite{Hagemann2021} if we view SNMs as a pair of Markov chains $((X_0, \dots, X_t),(Y_t, \dots, Y_0))$ where $(Y_t, \dots, Y_0)$ is the reverse Markov chain of $(X_0, \dots, X_t)$, we can view DNFs as a type of SDFs with specific forward and backward layers
\begin{align*}
\mathcal{K}_t(x,\cdot) &= P_{X_t|X_{t-1}=x} = \mathcal{N}(x+\epsilon g_{t-1}(x), \epsilon h^2_{t-1})\\
\mathcal{R}_t(x,\cdot) &= P_{Y_{t-1}|Y_{t}=x} = \mathcal{N}(x+\epsilon(g_{t}(x)-h^2_t s_t(x)), \epsilon h^2_{t})
\end{align*}
The equations above come from the Euler discretization with step size $\epsilon$ of the stochastic differential equation with drift $g_t$, diffusion coefficient $h_t$ and Brownian motion $B_t$
\[
dX_t = g_t(X_t)dt + h_t dB_t
\]

\section{Discussion}
In this section, we talk about the role of stochasticity in normalizing flows and compare the various techniques introduced above on the basis of the following criteria:
\begin{itemize}
    \item \textit{Expressivity}- while expressivity is usually used in a broad sense in the literature, we focus on each technique's ability to capture the various modes of the distribution they are trying to model as well as regions with relatively low density.
    \item \textit{Training speed}- we characterize training speed as the time taken by each technique to reach convergence.
    \item \textit{Ease of likelihood computation}- for this criterion, we look at the tractability of the likelihood computation for density estimation.
    \item \textit{Sampling efficiency}- we differentiate sampling efficiency from data efficiency, with the former referring to the computational cost required to generate samples with the latter referring to the number of samples required for optimization.
\end{itemize}

We also direct the reader to the comprehensive comparison of the bulk of the techniques covered in this paper performed by \citet{Bond_Taylor_2022}

\subsection{Expressivity}
As described in section \ref{sec:vae}, VAEs employ the use of latent variables. The choice of these provides them with a great deal of flexibility, resulting in highly expressive models. On the other hand, bijectivity constraints imposed by the normalizing flows framework result in representational insufficiency. Their representational capacity depends on the type of flow used in the model. For example, linear flows are limited in their expressivity. On the other hand, coupling and autoregressive flows, two of the most widely used flow architectures, enable normalizing flows to represent very complex distributions. However, they are still limited in their expressivity due to the invertibility constraint imposed by the framework (\citealt{Kobyzev_2021}).



Stochastic normalizing flows overcome some of these limitations by incorporating stochastic sampling blocks into the normalizing flow framework, thus improving representational capacity over deterministic flow architectures by overcoming topological constraints (\citealt{wu2020}). DiffFlow enjoys better expressivity than vanilla normalizing flows by overcoming their bijectivity constraints by adding noise. The aforementioned constraints prevent normalizing flows from expanding density support to the whole space when transforming complex distributions to a base distribution. As a result, DiffFlow can learn distributions with sharper boundaries (\citealt{zhang2021})

Score-based methods are notably flexible due to the fact that they are independent of the normalizing constant $Z_{\theta}$. This allows score-based methods to represent a more diverse set of models. Similar to other types of models, score-based methods are limited by the constraint that the dimension between their input and output must match. Otherwise, score-based models may take the form of any vector-valued function and thus are quite expressive (\citealt{wallach:19}). 

Surjective flows empirically demonstrate an ability to represent sharper boundaries than vanilla NFs, however, their methods are non-general and require prior knowledge of relevant symmetries in the target distribution (\citealt{nielsen2020}).

\subsection{Training speed}
Normalizing Flows are known to be inefficient and difficult to train due to invertibility constraints on transformations and as a consequence input and output spaces with the same dimension (\citealt{Bond_Taylor_2022}). By adding noise and bypassing strong bi-Lipschitz limitations, stochastic normalizing flows are easier to optimize. Moreover, adding stochastic layers is not computationally costly since they have linear computational complexity \cite{wu2020}. 

DiffFlow tends to train slower in comparison to other models. While certain aspects, such as a trainable forward function, help improve efficiency, DiffFlow ultimately relies on backpropagation, making it slow to train. On the other hand, VAEs reach convergence quite quickly. Due to the reparameterization trick proposed by \cite{kingma-original-vae}, VAEs can use SGD during optimization. 



Score-based models may struggle with training for low density regions, especially if the target distribution has multiple modes with a degree of separation. The model may then fail to converge in a reasonable time. As mentioned in section \ref{sec:score}, adding progressively more noise to the data in training can improve model convergence in such cases. 

\subsection{Ease of Likelihood Computation}
Normalizing flows benefit from having bijective and invertible (diffeomorphisms) transformations applied to base distributions, resulting in the ability to compute exact likelihoods \cite{Kobyzev_2021}. Adding noise to stochastic and diffusion normalizing flows increases expressivity over normalizing flows but at the cost of not being able to compute exact likelihoods. The parameters of a stochastic normalizing flow can be optimized by minimizing the KL divergence between the forward and backward path probabilities. This minimization makes use of the variational approximation, which precludes them from computing exact likelihoods \cite{wu2020}. Diffusion normalizing flows add noise in the forward stochastic differential equation. Consequently, they use the reparameterization trick proposed by \citet{kingma-original-vae} and thus we cannot compute exact likelihoods. To estimate likelihoods they use the marginals distribution equivalent SDE \cite{zhang2021}.



VAEs optimize the variational lower bound, an approximation of the log-likelihood we are trying to optimize and as a result, we cannot compute exact likelihoods. Importance sampling or Monte Carlo sampling techniques are used to compute the likelihood of data after training is completed \cite{kingma-welling-intro-to-vae}. Finally, score-based methods provide an avenue to compute exact likelihoods. This requires some manipulation of the equations and the introduction of invertibility into the model. According to \cite{score:likelihood}, score-based methods are then able to achieve `state-of-the-art likelihoods' on some image generation tasks.

Among the transformations proposed in \cite{nielsen2020}, only inference surjections, i.e. surjective layers that have full support in the base distribution and partial support in the target distribution, are able to produce exact likelihoods. Generative surjections, on the other hand, can only provide stochastic lower bound estimates.


\subsection{Sampling Efficiency}
Sampling efficiency is mainly affected by the complexity of the model and number of iterations to generate a sample. For example, VAEs consist of an encoder and decoder that are typically complex neural networks. On the other hand, VAEs can generate samples from a single network pass and are thus more efficient than other energy-based models such as stochastic normalizing flows \cite{Bond_Taylor_2022}.

The sampling efficiency of normalizing flows is related to the cost of the generative direction. However, since the transformations applied to the base distribution are deterministic, samples can be generated in a single network pass and thus, normalizing flows enjoy high sampling efficiency.  In comparison, diffusion normalizing flows have poor sampling efficiency, since they require MCMC during sampling. Nevertheless, they have better sampling efficiency than diffusion probabilistic models since they require fewer discretization steps \cite{zhang2021}. Similar to diffusion normalizing flows, stochastic normalizing flows have lower sampling efficiency than vanilla normalizing flows because they use an MCMC method, Metropolis-Hastings algorithm, to generate samples. 



Score-based methods tend to be slow in generating samples, due to the iterative nature of their sampling process. However, score-based methods are often able to produce high quality samples, comparable to GANs in image generation (\cite{wallach:19}).

\subsection{On the Role of Stochasticity}
Both Stochastic Normalizing Flows and Diffusion Normalizing Flows introduce stochasticity into their model formulations, though they provide different explanations for the role that stochasticity plays in improving expressivity. \cite{wu2020} frame the addition of stochastic layers in SNFs as incorporating the strengths of deterministic methods at performing large-scale probability transport with the fine-grained expressivity of Metropolis-Hasting MC sampling-- effectively removing samples in areas of lower probability density without incurring the sampling time costs of running a fully MC-reliant model (\citealt{wu2020}). \cite{zhang2021}, on their other hand, attribute the expressivity improvements by DNFs to an expansion of the training support to larger areas of the ambient space, improving gradient coverage for training (\citealt{zhang2021}). 

Both agree that adding stochasticity is central to bypassing topological constraints and representing sharp density boundaries in the target space, but the exact mechanism by which it improves expressivity is not fully elucidated by either work. Though beyond the scope of this paper, \cite{colddiffusion} demonstrates experimental evidence of successful diffusion-like models trained using deterministic, non-Gaussian forward processes, such as blurring and masking, calling into question the need for stochastic noise at all. None of the surjective layers proposed \cite{nielsen2020} utilize added noise, yet they are nonetheless able to represent sharp boundaries in the target distribution. The role and necessity of added noise in improving model expressivity are not clear from these works and require further investigation.

\section{Conclusion}
This paper delved into a variety of generative models and compared the relative performance of each on expressivity, training speed, sample efficiency and likelihood tractability. Starting from a basis of likelihood-based models, we explored the ability of Normalizing Flows and Variational Autoencoders to directly learn a distribution’s probability density in addition to their capacity to generate samples. VAEs have an encoder-decoder framework trained by optimizing the evidence lower bound, while NFs are structured as a series of bijective differentiable transformations on a data distribution to a simple base distribution. 

The strict constraints of the NF architecture narrow the types of distributions that the model can represent. Thus, we explored several models that relaxed the strict bijectivity constraint of NFs. The variations we studied borrowed aspects from different frameworks, including diffusion and score-based models, and introduced stochasticity into the training process. The introduction of noise adds both flexibility and stability to these models. The variations of NFs have performed well in practice, particularly on sampling tasks like image generation. While they cannot be used to compute exact likelihoods, they add much to the field in terms of expressivity and sampling efficiency. 
\newpage

\vskip 0.2in
\bibliography{main}

\end{document}